# Digital Twin-Assisted Controlling of AGVs in Flexible Manufacturing Environments

Mohammad Azangoo[1], Amir Taherkordi[2], Jan Olaf Blech[1, *], and Valeriy Vyatkin[1, 3]

[1]*Department of Electrical Engineering and Automation, Aalto University, Espoo, Finland*
[2]*Department of Informatics, University of Oslo, Oslo, Norway*
[3]*Department of Computer Science, Electrical and Space Engineering, Luleå University of Technology, Luleå, Sweden*
Emails: mohammad.azangoo@aalto.fi, amirhost@ifi.uio.no, jan.blech@aalto.fi, valeriy.vyatkin@aalto.fi

*Abstract*—Digital Twins are increasingly being introduced for smart manufacturing systems to improve the efficiency of the main disciplines of such systems. Formal techniques, such as graphs, are a common way of describing Digital Twin models, allowing broad types of tools to provide Digital Twin based services such as fault detection in production lines. Obtaining correct and complete formal Digital Twins of physical systems can be a complicated and time consuming process, particularly for manufacturing systems with plenty of physical objects and the associated manufacturing processes. Automatic generation of Digital Twins is an emerging research field and can reduce time and costs. In this paper, we focus on the generation of Digital Twins for flexible manufacturing systems with Automated Guided Vehicles (AGVs) on the factory floor. In particular, we propose an architectural framework and the associated design choices and software development tools that facilitate automatic generation of Digital Twins for AGVs. Specifically, the scope of the generated digital twins is controlling AGVs in the factory floor. To this end, we focus on different control levels of AGVs and utilize graph theory to generate the graph-based Digital Twin of the factory floor.

## I. INTRODUCTION

The manufacturing paradigm has been expanded towards flexibility to deal with diverse production demands. In this context, the Digital Twin concept helps achieve more agility and also interoperability between real manufacturing systems and their simulation models [1]. The Digital Twin concept has become popular in the context of Industry 4.0 and smart manufacturing; nowadays, many companies are aiming to improve their facilities through digitalization and using Digital Twins. They increase the competitiveness, productivity and efficiency of the main disciplines of manufacturing systems such as production planning, control, maintenance and layout planning [2]. Generally, a Digital Twin refers to the digital counterpart of a (usually) physical entity. In addition, some form of continuous interaction and communication between the physical entity and its digital counterpart is assumed. In industrial automation, examples of Digital Twins range from geometric models of a workpiece or a machine, to predictive models based on neural networks. In this context, we are primarily interested in Digital Twins that can be described using formal description mechanisms such as graphs. We believe that the use of formal descriptions will make the Digital Twin easier to be applied from different requirement perspectives, such as formal consistency analysis or fault detection.

The automatic generation of Digital Twins is an emerging research field and can save time and reduce costs. It can also respond to high demands from industries that want to digitize their production lines and systems [3]. For the generation of Digital Twins we may create a simulation model and specify its connection to the real system. In this way, any form of available information in the project life cycle can be used to accelerate the generation of Digital Twins in an automatic way. There exists some sporadic content in the academic literature about the automatic generation of Digital Twins, however it is very limited and also focused on specific issues. Thus, there is a need for more comprehensive approaches to automatic generation of Digital Twins in manufacturing systems.

Automated Guided Vehicles (AGVs) are considered a relevant domain in this context. They can move around the factory floor and extract plenty of information for automatic Digital Twin generation. Using AGVs is very common in flexible manufacturing environments, where the structure of the system can be reformed based on changes in customer or internal requirements. The mission for AGVs can be changed in less than a second to support a new mission on a factory floor.

As a general research direction in our Aalto Factory of the Future (AFoF) Laboratory[†], we aim to generate Digital Twins for flexible manufacturing environments to track workpieces.

By developing a chain of software tools we want to make this process as automatic as possible. To have a clearer research plan, it is divided to five consecutive steps:

1) *For each individual subsystem in the factory floor:* Extract the information required to generate the corresponding model.

---

[*]This paper is dedicated to the living memory of Professor Jan Olaf Blech who passed suddenly away on February 14, 2021.

978-1-7281-9023-5/21/$31.00 ©2021 IEEE

[†]https://www.aalto.fi/en/futurefactory

2) Generate a formal (graph) model of the system from the extracted information.
3) Make connections between the generated models and real systems to setup the Digital Twins.
4) Connect graph based Digital Twins for different subsystems to make a comprehensive Digital Twin for the whole factory floor.
5) Implement and run added-value services on Digital Twins.

It is assumed that the first step has been performed or that we can receive required information for this step in a proper format. We carried out steps 2 and 3 for the conveyor belt system in our last works [4], [5]. The goal of this paper is to cover these two steps again for the AGVs on the factory floor. The next steps, steps 4 and 5, are introduced for motivational purposes and they will be covered in the future research. However, some added-value services for generated Digital Twins have been used in this paper to show the benefits of generated graph-based Digital Twins.

This paper is organized as follows. The next section will review the literature on the automatic generation of Digital Twins and control systems for mobile robots. Section III introduces the layer-based architecture for AGV control on the factory floor and its relationship with Digital Twins. Section IV briefly presents the developed ideas about the automatic generation of graph-based models for factory floors and also explains the core idea of different control layers, implemented algorithms and modelling details with more supporting details. Section V discusses the current progress, initial results, and also the future plans to complete the work. Finally, Section VI concludes the paper.

## II. RELATED WORK

The automatic generation of Digital Twins has been recently reported in a number of research works in the smart factory domain. The goals for automatic Digital Twin generation in existing works are quite diverse, from digitization of old brownfield systems to paying less for engineering, and gaining more accuracy. In this section, we discuss related work in this area with a special focus on graph-based models for AGV control in manufacturing sites which serve as a key initial phase for obtaining efficient and accurate Digital Twin models in an automatic manner.

Software solutions that can generate Digital Twins automatically are very useful as they can save time and cost and improve the quality of Digital Twins. The authors in [6] propose an automatic solution for generating Digital Twins based on fast scanning and object recognition in the factory environment. Campos et al. [7] applied a high level specification to generate Digital Twins for industrial transportation and warehouse systems. Sierla et al. [8] extracted the steady state Digital Twin from Piping & Instrumentation Diagram (P&ID) documents in the brownfield process systems. Other than P&ID, the 3D scanned PCFs (Piping Component Files) can be used for automatic Digital Twin development [9]. At the moment, the tools for automatic generation of Digital Twins are very specific and limited; this paper presents part of a new software chain which can generate Digital Twins for manufacturing sites equipped with AGVs.

AGVs are equipped with different sensing technologies such as 2D and 3D scanning, lasers, lidars, accelerometers and encoders. These types of sensing equipment help AGVs comprehend more about their environments and can make a local map model for the surrounding environment to ease their operation. On the factory floors or in warehouses with several operational AGVs, the environment information can be gathered from all AGVs and infrastructures to create a comprehensive model. For example, Cardarelli et al. [10] present a cloud architecture for cooperative data fusion systems to create a live model of the industrial environment and support enhanced coordination of groups of AGVs. In [11], it is shown that using an active RFID system in the equipped fields can improve positioning of AGVs in their movements. In the current work, the benefits of using comprehensive live models generated from AGV information will be discussed.

A graph is made up of nodes which are connected by edges. In the engineering context, graphs are being used to model pairwise relations between objects. For example, in [12] graph theory has been used as a model for system dynamics to provide better formal analysis of dynamic complexity. Different subsystems of a complex system can also be modeled individually using graph theory. Then, it is possible to analyse the whole system by merging the individual graph models of each subsystems. Using graph-based modelling is very common for manufacturing plants, factory floors and AGV related systems. In [13], the environment map for AGVs in an automated warehouse is described by a graph model to make a collision free route planning system. Qiu et al. [14] in a survey on scheduling, path planning and routing algorithms for AGVs in flexible manufacturing environments listed several use cases and applications based on graph modelling which show the importance of this tool in this context. Since graph modelling is commonly used in the context of engineering systems, in this work we decided to use it to generate Digital Twins for factory floors.

Unified Modeling Language (UML) class diagrams can be used to show the structure of a system by using classes to describe entities of a system, inheritance and their relations to present relationships between classes. The Systems Modeling Language (SysML) provides a standard for describing system configurations in the automation industry. UML-based standards for describing configurations of systems have been established, e.g., in the automotive industry with the AUTOSAR standard. These can serve as a kind of Digital Twin, especially during development time [15]. In [5], we have shown the benefits of using UML class diagrams for modelling Digital Twins in flexible manufacturing environments. This paper uses a UML class diagram to show the big picture of the generated Digital Twin and its connection to the available neighboring systems. It is possible to generate a UML class diagram model for any flexible factory floor by merging UML class diagrams for different available subsystems on the factory floor.

## III. LAYER-BASED ARCHITECTURE FOR AGV CONTROL

To develop our idea about the layered-based architecture for AGV control, we reviewed similar concepts in Intelligent Transportation Systems (ITS) where the intelligent and connected sub-systems like vehicles, road infrastructures, data networks and local and cloud based control units are implemented to provide safer and more efficient road traffic streams. The California PATH program has designed a fully Automated Highway System (AHS) which introduces large-scale, hierarchical, hybrid control systems [16]. The network layer, which is the top layer in the control system of the AHS, uses a capacitated graph to control traffic flow within the AHS network. We believe that a similar architecture can be used in AGV control systems to deal with the complexity of flexible manufacturing.

In our work, we are porting these ideas to our Aalto Factory of the Future demonstrator. We identify entities such as AGVs, equipment and areas on the factory floor and identify their relationships. Some relationships have a dynamic nature such as an AGV moving on the factory floor from one area to another. Despite this, the identified entities and relationships should be applicable to a large class of industrial automation systems. As shown in Fig. 1, the UML class diagram has been used to show the architecture of layered-based control and Digital Twin for AGV automation on the flexible factory floor.

The *physical layer* contains all the on-board AGV controllers, fast safety actions and any other physic based components like motor speed and steering control systems. The main goal of this layer is to provide a precise local position, speed and rotation control. The *first control layer* is responsible for the design and execution of the maneuvers based on the current location of the AGV and its short term destination goal which comes from the second control layer. In this layer, for the sake of simplicity, the AGV is modeled as a particle (for example the center of mass of an AGV in space can be considered as the particle model for that AGV). The goal of this layer is to decide about continuous behaviour of the AGV. The role of the *second control layer* is path planning for AGVs in the graph model of the factory floor. This layer uses the general graph model of the factory floor and origin/destination matrix in order to optimize the total travel time for AGVs. The goal of this layer is to optimize high level cost functions using discrete mathematics tools.

From a high level perspective, it is clear that the physical layer represents the actual physical hardware components. However, the control layers can belong both to the physical control aspects of AGVs, as well as the Digital Twin models. In the proposed architecture, we adopt a combination of these two views of the control layers and propose to have a *hybrid Digital Twin* model. In such a model, based on the target application requirements, we can tune the level of physical and digital controls. Specifically, based on the type of added-value services defined in the system, we can expand the Digital Twin features by providing more control data and/or control functions at this level of abstraction, resulting in more interaction between the physical system and the counterpart Digital Twin. This may lead to more communication overhead, while it can offer better or more value-added services. From a more coarse-grained viewpoint, the system designer may decide to remove the first control layer from those modelled in the Digital Twin because of different reasons, such as not being able to digitally control or performing maneuvering in some AGV types.

As mentioned above, it is possible to implement some added-value services in Digital Twins in order to benefit more from the system [5]. Added-value services can improve the efficiency of systems and reduce the malfunction probability and safety risks. In [17], several added-value services and their benefits are listed for smart cars and transportation like autonomous navigation control, vehicle health monitoring, battery management systems, vehicle power electronics, and electrical power drive systems monitoring which can be used similarly in AGV related domains. Based on requirements from the project or customer, we can decide about the proper added-value services at the Digital Twin level.

## IV. DIGITAL TWIN AND CONTROL SYSTEM IMPLEMENTATION

In this section, details of implemented codes in the control layers and Digital Twin are discussed.

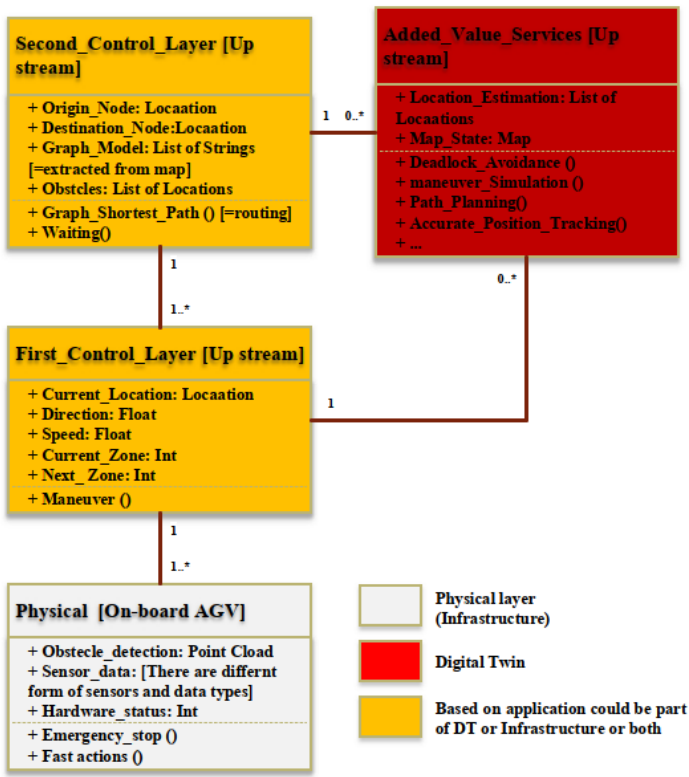

Fig. 1. layered-based UML class diagram for the control of AGVs in the factory floor.

## A. Generate models from extracted map images

The first step to generate a model of the factory floor is to extract the map information or 3D models of the environment from AGVs. Several studies have been carried out to generate precise map models of industrial environments. Beinschob et al. [18] introduces a semi-automated approach for the mapping an industrial environment using built-in sensors and scanners in AGVs for reducing the general installation costs and time. Recently, a research work was conducted at our laboratory [19] on factory floor model generation using AGVs. It shows the possibility of different solutions which AGVs can help to extract the information of the factory floor for generating corresponding models. Currently, two available AGVs in the AFoF, including MiR100 and Seit100, can generate a static map model of the factory floor in an image format, but it is necessary to convert these maps into a machine readable format in real-time. In the future, we aim to use image processing applications to make this step automatic as well.

## B. Using the graph theory to model the factory floor

If we consider a rectangle-shaped factory floor, then it can be divided into square-shaped zones. In cases where the shape of the factory floor is not a rectangle, the covering rectangle of that shape could be considered to cover the whole shape, then outside spaces can be marked as occupied. As shown in Fig. 2, a graph model can be generated from zones on the factory floor, presenting the connection between different zones on the factory floor. Two zones are connected if they have a common border line.

In the case of having $n*m$ zones, an initial adjacency matrix of the graph model can be defined as shown in Fig. 3.

The array elements in the adjacency matrix of the graph model shown in Fig. 3 are defined as below:

$$H_{abcd} = \begin{cases} 1 & \text{For neighbour zones} \\ & \text{when } (a = c \text{ and } b = d \pm 1) \\ & \text{or } (a = c \pm 1 \text{ and } b = d) \\ W & \text{Wait parameter} \\ & \text{when } a = c \text{ and } b = d \\ 0 & \text{For disconnected zones} \end{cases} \quad (1)$$

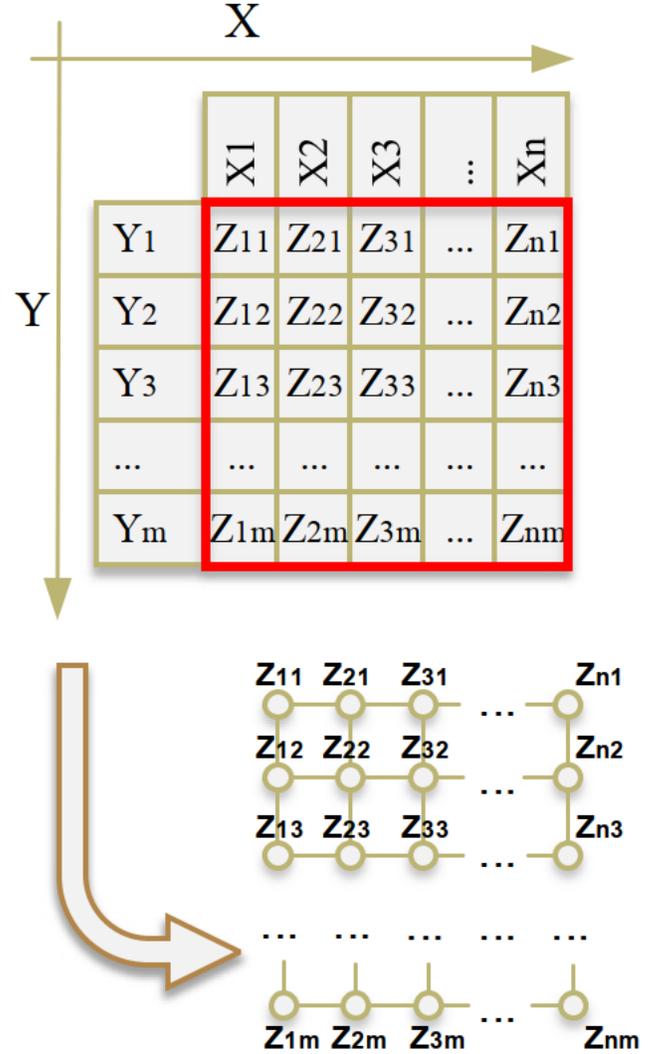

Fig. 2. Zones in the factory floor (top) and corresponding graph model (bottom).

The elements on the matrix diagonal represent self-transitions and can be used to show the required waiting time in case of a collision. If a zone like $Z_{ef}$ is occupied by an obstacle then all the corresponding elements in the adjacency matrix, i.e. all the items in the corresponding row and column for zone $Z_{ef}$, will be converted to zero.

## C. Zone transition in the second control layer

To evaluate our platform, we decided to implement simple control algorithms in different control layers and analyze the results for proof of concept. Digital Twin continually creates a new graph model from the latest maps extracted by AGVs and then the control algorithms can modify their results based on the updated model.

In the second control layer, Dijkstra's Shortest Path algorithm has been chosen to find the shortest path for each AGV from origin to destination. It is a simple Single-Input Single Output (SISO) algorithm and it cannot find the optimal solution for an environment with independent multi-agent systems. Because of the dynamic form of the environment, the recalculation for shortest path algorithms is required. The interval for recalculation can be defined based on size of the factory and density of humans and robots (dynamic obstacles).

On factory floors with several AGVs, the waiting strategy can enhance the routing result and reduce the collision probability [20]. In this strategy, first the zone before the conflict area should be found, then the low priority AGV needs to wait to let the high priority AGV pass that area. This is not

![Fig. 3. Adjacency Matrix.]

Fig. 3. Adjacency Matrix.

an optimal solution for multi-agent path planning, but it is simple for implementation and it works well for low traffic environments. Therefore, to make the SISO shortest path algorithm more efficient, we implement the simple waiting algorithm to avoid possible collisions and blocking. As shown in Fig. 4, the waiting algorithm can find the Collision Area and Waiting zones before the Collision Area for AGVs. The first AGV that can reach the Collision Area can continue moving in a normal form, but the other AGV needs to wait in the waiting zone to let the first AGV pass the Collision Area. Remaining AGVs will follow the same strategy to pass the collision area.

### D. Maneuvers in the first control layer

In this section, the trajectory/motion planning for AGVs in the first control layer will be discussed. To implement the discrete paths into continuous AGVs, it is required to consider manoeuvre regulations for different forms of transitions between zones. In Fig. 5, different forms of patterns for AGV manoeuvres have been depicted. In the presented square-shaped zones, straight and 90 degree turn manoeuvres are the only implemented patterns in the first control layer for the transitions of AGVs between zones.

To implement the first control layer, the state diagram is used. It also needs to receive zone information from the second control layer. In the state diagram the borders of the zones are considered as states and transition information can be found from current and next zone information which comes from the second control layer. The details of the implemented state diagram and algorithm are shown in Fig. 6.

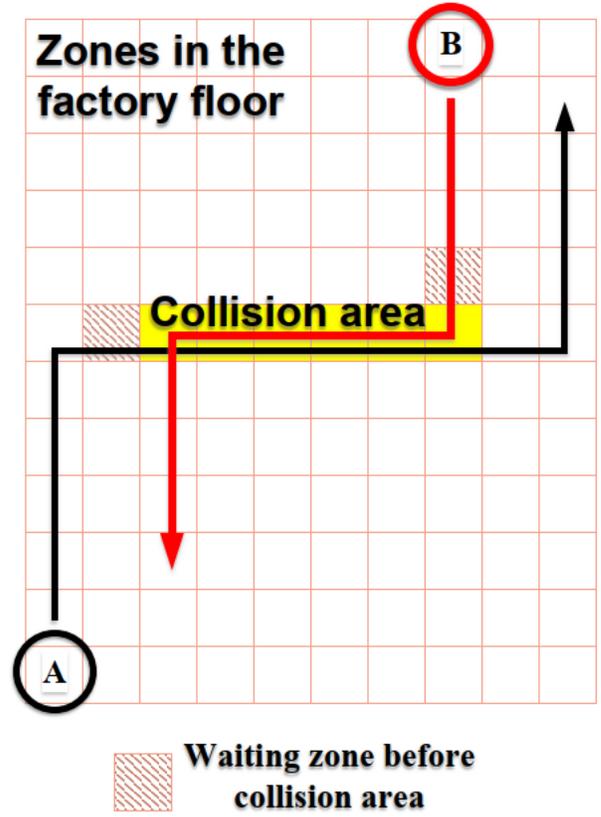

Fig. 4. A late AGV needs to wait in a waiting zone until the collision is resolved.

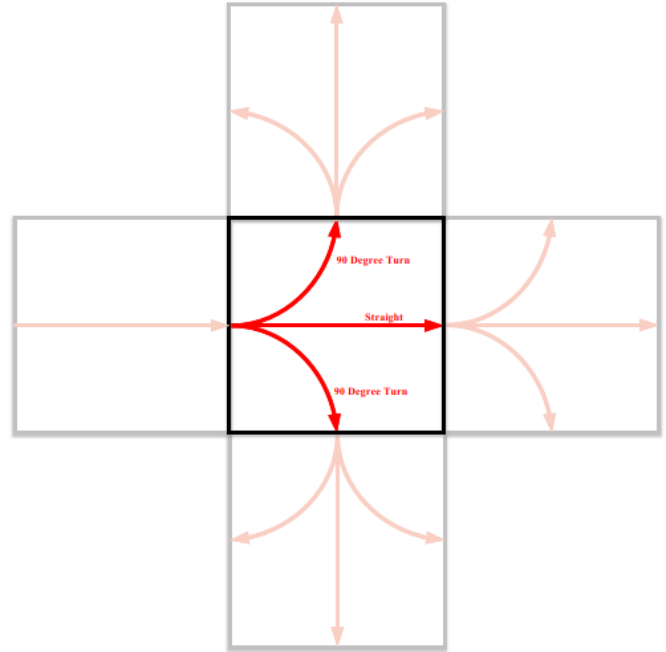

Fig. 5. 90 degree turn and straight maneuvers in the intermediate zones.

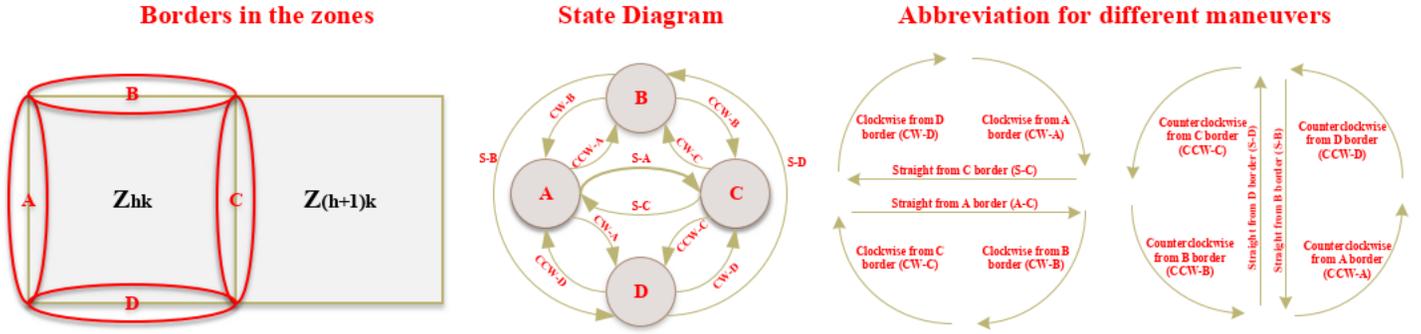

Fig. 6. (Left) Four border in the zones (Center) state diagram for transitions between borders (Right) the list of all possible maneuvers.

*E. Digital Twin*

In the current hybrid architecture, the control layers can be inside or outside the Digital Twin. Moreover, rather than direct control applications, several added-value services can be implemented in the Digital Twin to meet the requirements or improve the efficiency of the system. Added-value services can be in contact with different layers in the control system. In the current work, a simple planning service for charging batteries has been implemented in the Digital Twin for proof of concept. This service can count the covered distance and time for each individual AGV. Then, it can assign a new charging mission through communication with the second control layer. The requested AGV moves to a defined charging station and remains there for a specified time to charge its battery and then continue its work.

## V. Discussion

In this paper, a Digital Twin based framework is introduced for control of the AGVs at manufacturing sites. It is part of the main plan in AFoF for the automatic generation of Digital Twins in manufacturing. In this work, the general framework for control of AGVs and the rules of Digital Twins have been discussed. The main output of this research is a software tool which can be used for controlling AGVs at manufacturing sites with the help of Digital Twins. Our developed system is capable of enhancing the control and monitoring AGV systems in manufacturing environments. In the following paragraphs advantages and disadvantages of the developed methodology and also current implementation of the system will be discussed.

In this study, the graph model of the factory floor for controlling and monitoring the AGVs has been developed and generated. The graph modelling approach presented in this work for factory floors can be matched with different control layers and also Digital Twins. It is easy to develop, change and generate the graph-based models. There also exist several graph-based optimization algorithms which can be used at different levels of control hierarchy. However, it is important to work on modelling with higher fidelity in the future. For example, the attributes like general geographical direction and position of factory floors can be considered in the models for future integration.

In the current work, we only used simple solutions for the first and second control layers. However, they can be replaced by more optimal solutions. Moreover, new added-value services can be implemented in the Digital Twin to help control layers act more efficiently. For example, we used a simple SISO shortest path finding algorithm and also a simple Waiting algorithm for collision avoidance in the second control layer. The results showed that the system does not act optimally and in a crowded environment the probability of deadlock is high. In our future work, it is necessary to implement more optimal algorithms and added-value services to evaluate the efficiency of the proposed structure. In addition, in the first control layer, the current implemented maneuvers are very limited and can not translate all the actual requirements from industrial environments. The strength of our proposed framework is that it is made up of independent blocks, thereby it is possible to replace individual applications, control algorithms, and added-value services in the Digital Twin.

In general, the proposed structure can control the behaviour of AGVs at different levels. It can also empower the operation of flexible manufacturing systems with the help of Digital Twins, thereby it is a perfect starting point for future development. However, the level of fidelity was low in this work and it can be improved by considering the diversity of equipment on the factory floor (like different forms of obstacles: dynamic or static) and also physical parameters in models (for example the congestion or travel time can be added to graph models as a weight for edges).

A *sample recorded video for the operation* of the developed simulation model, multi-layer control and Digital Twin is available in [21]. The screenshot of the graphical user interface for monitoring the AGVs on factory floors are shown in Fig. 7.

## VI. Conclusion

In this paper, we proposed an approach towards automatic generation of Digital Twins in smart manufacturing systems, in particular for controlling AGVs in such systems. To achieve that, we first presented an architectural model of the control mechanism of AGVs, which encompasses both fine-grained and coarse-grained control levels. The former is concerned with the maneuvering of AGVs in their current location, while the latter is related to path planning for AGVs on the factory floor.

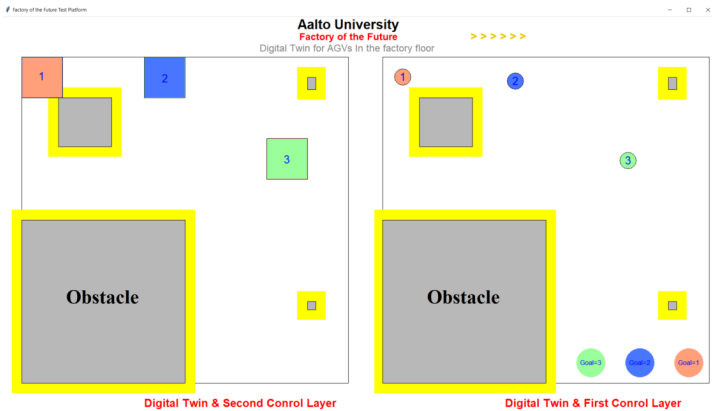

Fig. 7. The GUI which shows the behaviour of developed control layers and Digital Twin

Using graph theory, we proposed a mechanism for graph-based modeling of Digital Twins for AGVs, considering the aforementioned control requirements. In the proposed platform, interaction between the control layers and the Digital Twin can enhance the functionality of the system. The independent design of the control layers and added-value services in the Digital Twin ease the development of sub-systems.

Some developments and experiments are left for the future work such as implementing more advanced algorithms in different control layers and the Digital Twin side. In addition, development of automatic tools for generation and synchronization of the factory model should be considered. In addition, to achieve a Digital Twin with higher fidelity, it is important to study more detailed attributes for modelling.